\documentclass[letterpaper, 10 pt, conference]{ieeeconf}  %

\IEEEoverridecommandlockouts                              %

\usepackage{amsmath,amsfonts}
\usepackage{algpseudocode}
\usepackage[vlined,ruled]{algorithm2e}
\usepackage{array}
\usepackage{textcomp}
\usepackage{stfloats}
\usepackage{url}
\usepackage{verbatim}
\usepackage{graphicx}
\usepackage[outdir=./]{epstopdf}
\usepackage{cite}
\usepackage{mathtools}
\usepackage{float}
\usepackage{graphicx}
\usepackage{amsmath}
\usepackage{arydshln}
\usepackage{tensor}
\usepackage{multirow}
\usepackage{booktabs}
\usepackage{gensymb}
\usepackage{amssymb}
\usepackage{bbding}
\usepackage{bm}
\usepackage{subcaption}
\usepackage{siunitx}
\usepackage[colorlinks,linkcolor=black]{hyperref}
\usepackage[switch]{lineno}
\usepackage{relsize}
\usepackage{makecell}

\usepackage{xcolor}  %
\definecolor{bevtopo_color}{RGB}{255,0,0}
\definecolor{NetVLAD_color}{RGB}{164,217,187}
\definecolor{LoST_color}{RGB}{239,129,131}
\definecolor{R2Former_color}{RGB}{105,158,212}

\newcommand{\fref}[1]{Fig.~\ref{#1}}
\newcommand{\sref}[1]{Sec.~\ref{#1}}
\newcommand{\tref}[1]{Table~\ref{#1}}

\title{\LARGE \bf
SSMG-Nav: Enhancing Lifelong Object Navigation \\ with Semantic Skeleton Memory Graph
}

\author{Haochen Niu, Lantao Zhang, Xingwu Ji, Rendong Ying, Peilin Liu$^\dagger$ and Fei Wen%
\thanks{
This work was supported by the National Natural Science Foundation of China (No. 62276166) and the STI 2030-Major Projects (No. 2022ZD0208700).} %
\thanks{All authors are with the Brain-Inspired Application Technology Center (BATC), Shanghai Jiao Tong University, Shanghai 200240, China (email: \{haochen\_niu, swagger, jixingwu, rdying, liupeilin, wenfei\}@sjtu.edu.cn).}%
}

\begin{document}

\maketitle
\thispagestyle{empty}
\pagestyle{empty}

\begin{abstract}
Navigating to out-of-sight targets from human instructions in unfamiliar environments is a core capability for service robots. Despite substantial progress, most approaches underutilize reusable, persistent memory, constraining performance in lifelong settings.
Many are additionally limited to single-modality inputs and employ myopic greedy policies, which often induce inefficient back-and-forth maneuvers (BFMs).
To address such limitations, we introduce SSMG-Nav, a framework for object navigation built on a \textit{Semantic Skeleton Memory Graph} (SSMG) that consolidates past observations into a spatially aligned, persistent memory anchored by topological keypoints (e.g., junctions, room centers). SSMG clusters nearby entities into subgraphs, unifying entity- and space-level semantics to yield a compact set of candidate destinations.
To support multimodal targets (images, objects, and text), we integrate a vision-language model (VLM). 
For each subgraph, a multimodal prompt synthesized from memory guides the VLM to infer a target belief over destinations.
A long-horizon planner then trades off this belief against traversability costs to produce a visit sequence that minimizes expected path length, thereby reducing backtracking.
Extensive experiments on challenging lifelong benchmarks and standard ObjectNav benchmarks demonstrate that, compared to strong baselines, our method achieves higher success rates and greater path efficiency, validating the effectiveness of SSMG-Nav.

\end{abstract}

\section{INTRODUCTION}\label{sec:intro}
Service robots are expected to navigate to target objects beyond their current field of view in previously unseen environments based on human instructions. 
Conventional methods typically rely on reinforcement learning (RL) or imitation learning (IL) \cite{lu2025MultiView, ramrakhya2022HabitatWeb}. Despite notable progress, these methods exhibit two fundamental limitations: (i) they are data- and interaction-intensive, making training costly; and (ii) they often degrade under distribution shifts when deployment scenarios diverge from the training domain.

Human navigation suggests prioritizing high-likelihood locations (e.g., bedrooms or shelves for a stuffed toy), highlighting the value of commonsense spatial semantic reasoning. 
Meanwhile, large-scale pretrained models have shown strong generalization in commonsense and reasoning, offering a new paradigm for navigation.
In recent years, numerous zero-shot object navigation methods that draw on vision-language models (VLMs) or large language models (LLMs) have emerged, demonstrating promising advantages by leveraging broad prior knowledge to link environmental semantic cues with target objects \cite{gadre2023CoWs,yokoyama2024VLFM,long2024InstructNav,yin2025UniGoal}.

Nevertheless, several gaps remain for real-world, user-centered navigation, especially in lifelong settings where agents are expected to leverage historical experience for continual improvement \cite{chang2023GOAT, khanna2024goatbench}.
First, the design of memory mechanisms remains underexplored: many methods make purely reactive decisions from instantaneous observations; others construct value maps\cite{yokoyama2024VLFM,ramakrishnan2022PONI} or semantic maps\cite{zhou2023ESC,kuang2024OpenFMNava} that lack persistence and reusability. Recent efforts on graph memory are promising but most of them model “objects”\cite{yin2024SGNav,yin2025UniGoal} or "waypoints"\cite{zhan2024MCGPT} as nodes, connect them via adjacency relations, and construct the memory graph in a text-based form, which limits the overall expressiveness.
Second, most methods target single-modality goals and cannot accommodate multimodal instructions. 
Third, planners are commonly greedy, driving the agent to the most likely target locations with short-sighted decisions. This often leads to inefficient back-and-forth maneuvers (BFMs)\cite{niuSkeletonBased}.

These limitations motivate a navigation framework that maintains persistent memory and supports multimodal goals for efficient lifelong task execution. At its core is a Semantic Skeleton Memory Graph (SSMG) that provides reusable, spatially grounded semantics to support long-horizon planning.
Unlike existing graph representations that centers on objects\cite{yin2025UniGoal} or waypoints\cite{zhan2024MCGPT} as nodes, SSMG aggregates objects around topological keypoints of the environment skeleton (e.g., corridor junctions, room centers). This yields a set of candidate destinations, each represented as a subgraph that naturally combines entity semantics and spatial semantics.
Then we incorporate a VLM and, for each SSMG subgraph, derive a multimodal prompt to guide the VLM in evaluating belief that the target lies in that subgraph.
Leveraging the VLM’s multimodal commonsense and reasoning, we can accommodate targets specified as images, objects, or text while fully exploiting the local scene context encoded by each subgraph.
Moreover, we design a long-horizon planner that jointly considers belief and traversability cost to compute a visit sequence minimizing the expected travel distance. 
Using an efficient approximate solver to address this optimization, the planner effectively mitigates myopic behavior and improves navigation efficiency.

The main contributions are as follows.
\begin{itemize}
    \item 
    A novel Semantic Skeleton Memory Graph (SSMG) that consolidates historical observations into persistent, spatially grounded memory, unifying entity and spatial semantics to support informed decision-making.
    
    \item 
    An SSMG-driven long-horizon planner that elicits VLM-based environmental reasoning via multimodal prompts and employs an efficient approximation to balance belief and traversal cost for long-term visitation planning, improving efficiency.
    
    \item
    A complete lifelong multimodal object navigation framework, SSMG-Nav, which achieves state-of-the-art performance on public benchmarks.

\end{itemize}

\section{RELATED WORK}\label{sec:relatedwork}

\subsection{Object Navigation}
Existing approaches to object navigation can be broadly categorized into learning-based and zero-shot methods.
\subsubsection{Learning-based Object Navigation}
Learning-based methods have progressed rapidly, with reinforcement learning (RL) and imitation learning (IL) taking center stage. Building on large-scale human demonstrations, Habitat-Web \cite{ramrakhya2022HabitatWeb} shows that IL can internalize the implicit, often intricate behaviors underlying object search. On the modular front, PONI \cite{ramakrishnan2022PONI} uses a potential-function network to decide “where to look?” and pairs it with an analytical local policy for action generation, while SGM \cite{zhang2024Imagine} complements this line by predicting maps beyond the agent’s visible field via self-supervision. ZSON \cite{majumdar2022zson} brings zero-shot generality into the picture by training a policy with DD-PPO and leveraging CLIP \cite{radford2021Learning} for image-goal navigation. More recently, MVVT \cite{lu2025MultiView} taps multi-view inputs to couple global image cues with local spatial context, further boosting policy learning. 
Despite these advances, learning-based methods typically require large-scale data and tend to degrade under distribution shift.
\subsubsection{Zero-shot Object Navigation}
In recent years, powered by advances in large models, zero-shot object navigation has seen a surge of new methods.
Some studies leverage vision-language models (VLMs), such as CLIP\cite{radford2021Learning}, to align observations with target objects, incorporating advanced exploration or learning-based strategies for open-set object navigation \cite{gadre2023CoWs}. However, these methods lack analysis of environmental semantics and memory, resulting in lower exploration efficiency.

PIVOT \cite{nasiriany2024PIVOT} and VLMNav \cite{goetting2025EndtoEnd} cast navigation as video question answering, querying a VLM to select motion directions. 
Building on this idea, DyNaVLM \cite{ji2025DyNaVLM} augments the system with a graph memory architecture, representing objects as nodes and relational adjacencies as edges.

Methods such as PONI \cite{ramakrishnan2022PONI}, VLFM \cite{yokoyama2024VLFM} and WMNav \cite{nie2025WMNav} construct a value map to score frontier points on the map, thereby guiding the agent to prioritize exploration of regions with higher target relevance.
However, such value maps are often not reusable because they are tightly coupled to the current task.

In addition, several approaches construct a semantic map. Methods like ESC \cite{zhou2023ESC} L3MVN \cite{yu2023L3MVNa}, and OpenFMNav \cite{kuang2024OpenFMNava} aggregate object categories, locations, and relations into a semantic map and render candidate waypoints in the egocentric view as prompts; an LLM then chooses the next waypoint. 
Extending this paradigm, InstructNav \cite{long2024InstructNav} leverages dynamic chain-of-navigation (DCoN) and spatiotemporal chains \cite{wei2023ChainofThought} to explicitly structure spatial reasoning in natural language.

Some recent work focuses on graph representations of memory/map that can efficiently integrate different kinds of information. 
SG-NAV \cite{yin2024SGNav} constructed a hierarchical scene graph \cite{gu2024ConceptGraphs} and proposed a CoT prompting method for LLMs to reason about target locations.
Similarly, SayNav \cite{rajvanshi2024SayNav} created an object-room-scene topological map using objects as nodes in a bottom-up way. 
MCGPT \cite{zhan2024MCGPT} samples viewpoints during movement as topological nodes and utilizes an offline constructed CoT database to retrieve the most suitable prompt templates for online analysis.

\subsection{Lifelong Multimodal Object Navigation}
Recently, some researchers have begun to focus on object navigation tasks that better align with real-world scenarios. 
\cite{yin2025UniGoal} pointed out the limitations of existing zero-shot methods in constructing reasoning frameworks on LLMs for specific tasks, noting that they cannot generalize across different types of targets. 
Therefore, they proposed a graph structure with objects as nodes for unified multimodal target representation, verifying targets through multi-stage subgraph matching based on LLM inferences, though lifelong task execution was not considered.
SayNav \cite{rajvanshi2024SayNav} studied multi-object navigation tasks but can only present all targets at once and operates in a single modality. 
OpenIN \cite{tang2025OpenIN} focuses on lifelong task execution but is also limited to a single modality.
In contrast, GOAT \cite{chang2023GOAT} formalizes Lifelong Multimodal Object Navigation with sequential goals specified by object categories, images, or language descriptions, and GOAT-Bench \cite{khanna2024goatbench} provides a recent standardized benchmark with a unified protocol and public baselines for fair comparison.

\section{METHOD}
\begin{figure*}[!t]
    \centering
    \subfloat{
        \includegraphics[width=0.95\linewidth]{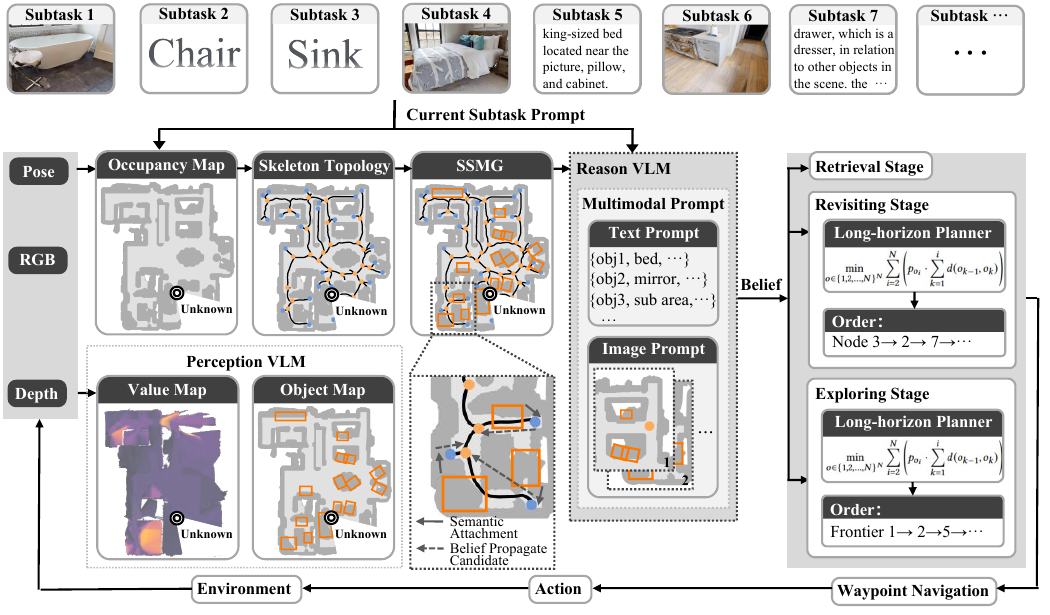}}
    \caption{An overview of the proposed SSMG-Nav.}
    \label{fig:system}
\end{figure*}

\subsection{Task Definition and System Overview}
The lifelong multimodal object navigation task places an agent in a previously unseen scene and requires it to navigate sequentially to multiple goal objects, each specified by an object category (e.g., “towel”), a text description (e.g., “a bath towel hanging on a hook in the bathroom, relative to the shower stall and shower bench”), or a unique image.
Each goal defines a subtask, which ends when the time budget is exhausted or the agent issues a “stop” action, after which it switches to the next subtask until completion. At each switch, the subtask is successful if the agent is within a distance threshold of the goal object; otherwise, it fails.

As shown in \fref{fig:system}, our proposed SSMG-Nav is structured as follows.
At each step, the agent receives RGB, depth, and pose from the simulator. Using these sensor inputs, we incrementally update geometric information (occupancy map and skeleton topology) and semantic information (value map and object map), thereby constructing the SSMG over time (\sref{sec:ssmg}). The navigation policy then transitions between states (\sref{subsec:policy}). Conditioned on the current state, we generate multimodal prompts from SSMG subgraphs to guide a VLM in belief estimation (\sref{subsubsec:llm}). Finally, a long-horizon planner jointly optimizes traversal cost and belief to produce a long-term visitation route and executes its first waypoint (\sref{subsubsec:planning}).

\subsection{Semantic Skeleton Memory Graph Construction}\label{sec:ssmg}

\subsubsection{Visual Perception}  
At each time step \(t\), the agent acquires sensor observations \({obs}_t = \{I_t, D_t, P_t\}\), comprising the RGB \(I_t\), depth \(D_t\), and pose \(P_t\).

We first convert the depth \(D_t\) into a 3D point cloud and project it onto a bird’s-eye-view (BEV) plane. Using ray casting from the sensor origin, we construct and incrementally update a 2D occupancy grid \(M_{\text{occ}}\). Each grid cell is labeled as \emph{free}, \emph{occupied}, or \emph{unknown} to track exploration progress.

To analyze the semantic information in the observations, we employ BLIP-2 \cite{li2023BLIP2} to evaluate the relevance between the current observation and the target. Following \cite{yokoyama2024VLFM}, we adopt the text template
\emph{``Seems like there is a \textless target object\textgreater~ ahead.''}
to perform text–image matching on the current RGB input \(I_t\).
Analogous to \(M_{\text{occ}}\), we project the relevance scores onto the BEV plane to obtain a value map \(M_{\text{value}}\), which encodes the correlation between the scene context and the target object across the explored area.

While the value map captures target-specific relevance, lifelong navigation requires a target-agnostic semantic representation of observations. 
We therefore use BLIP-2 again to generate a caption for the current image \(I_t\), denoted as \(\texttt{cap}_t\).
To obtain structured semantics, we apply the open-set detector Grounding-DINO~\cite{gdino} to \(I_t\) and produce a set of detections
\begin{equation}\label{eq:detected_objects}
  \mathrm{Det}_t = \{(\mathbf{b}_i, s_i, c_i) \mid i = 1,\ldots,N\},
\end{equation}
where \(\mathbf{b}_i\) is the bounding box, \(s_i\) the confidence score, \(c_i\) the predicted category, and \(N\) the number of detected objects.
For each detection in \(\mathrm{Det}_t\), we apply MobileSAM~\cite{mobile_sam} within the corresponding bounding box to obtain an instance mask. 
By fusing the mask with the depth, we back-project pixels to 3D and construct an object-level point cloud map \(M_{\text{obj}}\).
For each object \(j \in \mathcal{O}\), we store:
\begin{equation}\label{eq:object_info}
  \mathcal{O}_j := \{\mathbf{PC}_j,\, c_j,\, \mathcal{V}_j\},\quad
  \mathcal{V}_j := \{(\mathbf{I}_j^{k}, \mathbf{P}_j^{k}, \texttt{cap}_j^{k}, s_j^{k})\}_{k=1}^{K}.
\end{equation}
Here \(\mathbf{PC}_j\) is the point cloud, \(c_j\) the category, and \(\mathcal{V}_j\) collects up to \(K\) views: RGB \(\mathbf{I}_j^{k}\), pose \(\mathbf{P}_j^{k}\), caption \(\texttt{cap}_j^{k}\), and score \(s_j^{k}\). We set \(K=3\) in our experiments.

Over time, \(M_{\text{occ}}\) and \(M_{\text{value}}\) are incrementally updated via coverage. 
Updates to \(M_{\text{obj}}\) are gated by semantic category consistency and the alignment between 3D bounding boxes of the point cloud.
For objects subject to fusion, historical entries \(\mathcal{V}_j\) are replaced based on detection confidence, thereby retaining higher-confidence observations that typically correspond to better viewpoints. 

\subsubsection{Skeleton Topology Extraction}
Object-cluster graphs \cite{yin2024SGNav} may lose geometric fidelity, while trajectory-based node generation \cite{zhao2024OVERNAV} can yield non-representative nodes. 
We therefore aim to extract a concise geometric backbone and enrich it with semantics later.

Following \cite{niuSkeletonBased}, we approximate the skeleton of \(M_{\text{occ}}\) via morphological thinning to obtain a one-pixel-wide medial axis. 
Each skeleton pixel is classified as endpoint  \(\mathcal{N}_{\mathrm{e}}\) (degree 1), connector \(\mathcal{N}_{\mathrm{c}}\) (degree 2), or junction \(\mathcal{N}_{\mathrm{j}}\) (degree \(\ge 3\)). 
We then prune connectors and retain only \(\mathcal{N}_{\mathrm{e}}\) and \(\mathcal{N}_{\mathrm{j}}\) (special nodes), adding edges between them. 
This yields the graph \(\mathcal{G}=(\mathcal{N},\mathcal{E})\), where connectors are stored as edge attributes to support fast path planning.

\subsubsection{Semantic Attachment}
We augment the skeletal topology \(\mathcal{G}=(\mathcal{N},\mathcal{E})\) with object information from \(M_{\text{obj}}\) to obtain a complete SSMG: for each object \(j \in \mathcal{O}\), we select its highest-confidence viewpoint \(k_j^\star := \arg\max_{k} s_j^{k}\) and set the reference pose \(\mathbf{P}_j^\star := \mathbf{P}_j^{k_j^\star}\); projecting pose to BEV via \(\pi(\cdot)\), we attach the object to its nearest node:
\begin{equation}
n_j^\star := \arg\min_{n \in \mathcal{N}} \big\| \mathbf{x}_n - \pi(\mathbf{P}_j^\star) \big\|,
\mathcal{O}(n_j^\star) \leftarrow \mathcal{O}(n_j^\star) \cup \{j\}.
\end{equation}
To improve coverage while remaining concise, each node \(n\) defines a data-driven radius from the farthest currently attached object,
\begin{equation}
r_n :=
\begin{cases}
\max\limits_{j \in \mathcal{O}(n)} \big\| \mathbf{x}_n - \pi(\mathbf{P}_j^\star) \big\|, & \mathcal{O}(n)\neq \emptyset,\\[2pt]
0, & \text{otherwise},
\end{cases}
\end{equation}
and adds all objects within this radius,
\begin{equation}
\mathcal{O}(n) \leftarrow \mathcal{O}(n) \cup 
\Big\{\, j \in \mathcal{O} \ \big|\ \big\| \mathbf{x}_n - \pi(\mathbf{P}_j^\star) \big\| \le r_n \Big\},
\end{equation}
which naturally allows objects to belong to multiple nodes (i.e., \(j \in \mathcal{O}(n_1)\cap \mathcal{O}(n_2)\)). 
To support downstream belief propagation, we augment the edge set by linking each endpoint to directly connected junction node(s) on the skeleton (i.e., one-hop connectivity on the skeleton graph):
\begin{equation}
\mathcal{E} \leftarrow \mathcal{E} \cup 
\big\{\, (n_e,n_j) \in \mathcal{N}_{\mathrm{e}} \times \mathcal{N}_{\mathrm{j}} \ \big|\ \text{skeleton\_adj}(n_e,n_j) \,\big\},
\end{equation}
where \(\text{skeleton\_adj}(n_e,n_j)\) is true iff \(n_e\) and \(n_j\) are one-hop neighbors on the skeleton topology.

\subsection{SSMG-driven Long-Horizon Planner}\label{subsec:planner}

Zero-shot object navigation increasingly leverages VLMs/LLMs to translate perception into planning priors. 
However, prevailing systems select the single most promising frontier at each step \cite{yin2024SGNav,yin2025UniGoal,zhou2023ESC}, without sequencing multiple candidates over long horizons, which frequently results in BFMs and inefficient trajectories—a phenomenon also reported in autonomous exploration \cite{niuSkeletonBased}. 
Moreover, limitations in spatial understanding \cite{chen2024SpatialVLM} and long-horizon planning \cite{aghzal2025Survey,zhang2024FLTRNN}  cast doubt on the reliability of directly feeding large-scale global maps and rich semantic information into LLMs in some form for long-horizon planning.
We therefore propose an SSMG-driven Long-Horizon Planner (LHP) that builds SSMG-based multimodal prompts for reasoning over nodes of SSMG, followed by an approximation algorithm for long-term path planning.

\subsubsection{VLM-Based Belief Evaluation}\label{subsubsec:llm}
Belief estimation is performed per special node. We first build multimodal prompts from endpoints\(\mathcal{N}_{\mathrm{e}}\). Each node stores nearby object information; for each object \(j\in\mathcal{O}(n_i)\), we form a textual prompt \(\{ \text{object\_id}_j, \text{object\_class}_j, \text{object caption}_j \}\). We then extract a subgraph as an image prompt: centered at node \(n_i\), we annotate the BEV plane with each object's point-cloud centroid and its \(\text{object\_id}_j\). Concatenating the textual and visual prompts yields a background-knowledge prompt. Finally, CoT reasoning guides the VLM (Qwen-VL-Plus\cite{Qwen-VL}) to analyze node \(n_i\) step by step and produce a formatted belief estimate, $\mathit{belief}_i$.

After evaluating all endpoints \(\mathcal{N}_{\mathrm{e}}\), we propagate endpoint cues bottom-up to \(\mathcal{N}_{\mathrm{j}}\). This focuses candidates on junctions, reducing the planning search space. For each junction, we augment its background-knowledge prompt with information from its associated endpoints and query the VLM again to output the junction belief.

\subsubsection{Long-Horizon Planning}\label{subsubsec:planning}
Our LHP treats all \( \mathcal{N}_{\mathrm{j}} \) as candidate nodes and computes a visitation path that sequentially accesses them. The agent proceeds along this path until the target object is found.
Given \(N\) candidate nodes \(\{n_1,\ldots,n_N\}\) with target probabilities \(\{p_1,\ldots,p_N\}\) and pairwise distances \(d(i,j)\), we seek an order \(\{n_{o_1},\ldots,n_{o_N}\}\) that minimizes the expected travel while searching for the target:

\begin{equation}\label{eq:2opt}
    \min_{o \in \{1, 2, \ldots, N\}^N} \sum_{i=2}^{N} \left( p_{o_i} \cdot \sum_{k=2}^{i} d(o_{k-1}, o_k) \right),
\end{equation}
subject to:
\begin{equation}\label{eq:cons}
    \begin{aligned}
        \sum_{i=1}^{N} p_i = 1, \quad p_i \geq 0, \\
        | \{ o_1, o_2, \ldots, o_N \} | = N.
    \end{aligned}
\end{equation}

This problem is NP-hard. Inspired by the 2-opt local search heuristic \cite{croes1958Method}, we propose an approximate solution. We first convert the estimated \( \text{belief}_i \) from \sref{subsubsec:llm} into probabilities \( p_i \) via softmax with temperature $\tau$:

\begin{equation}
    p_i = \frac{\exp\!\big({belief}_i/\tau\big)}{\sum_{j=1}^{N} \exp\!\big({belief}_j/\tau\big)}, \quad \tau > 0.
\end{equation}

We first obtain an initial path via a greedy heuristic that always visits the nearest unvisited node. For simplicity, let \(n_1\) denote \(n_{o_1}\), and denote the initial path as \(\pi_0=\{n_1,n_2,\ldots,n_N\}\). Its expected travel cost \(E(\pi_0)\) is computed using \eqref{eq:2opt}.
Next, select a segment \(\{n_p,n_{p+1},\ldots,n_q\}\), reverse it, and replace the original segment to form \(\pi_{pq}=\{n_1,\ldots,n_{p-1}, n_q,n_{q-1},\ldots,n_{p+1},n_p,\ldots,n_N\}\).
Compute \(E(\pi_{pq})\); if it is lower than the current expectation, accept \(\pi_{pq}\). Iterate over all \((p,q)\) pairs and repeat until reaching the maximum iterations or no further improvement.

\subsection{Navigation Policy}\label{subsec:policy}
Based on the aforementioned content, we develop a strategy with four stages: retrieval, revisiting, exploring, and waypoint navigation.

\subsubsection{Retrieval Stage}
When switching to a new subtask, we first query the SSMG for the specified target. If same-category instances are found, we construct prompts as in \sref{subsubsec:llm}. If an instance is verified as the target, we proceed to waypoint navigation. If the target is absent from the current SSMG, we assess exploration status on \(M_{occ}\). If valid frontiers remain, we enter the exploring stage to search unexplored regions. If exploration is complete, we consider potential missed detections (e.g., due to viewpoint) and transition to revisiting.

\subsubsection{Revisiting Stage}
In this stage, we follow \sref{subsec:planner} to compute the access sequence \(\pi\) and visit nodes sequentially. At each junction, the agent performs a 360\(^\circ\) spin to acquire panoramic observations. If the target is matched, we transition to waypoint navigation; otherwise, we check for adjacent end nodes, visit them (if any), and repeat the 360\(^\circ\) scan. If still unmatched, proceed to the next junction in \(\pi\) and repeat.

\subsubsection{Exploring Stage}
In this stage, we also conduct long-horizon planning to improve exploration efficiency. Specifically, we extract the boundaries between known and unknown regions as frontiers. These frontiers are then clustered to obtain several frontier points $\{f_1, f_2, \ldots, f_n\}$. 
We then solve the path using the method described in \sref{subsubsec:planning}. Considering the incompleteness of SSMG information and the uncertainty of unknown areas during the exploration process, we simply use a value map \(M_{\text{value}}\) as the belief for each frontier point, similar to the approach in \cite{yokoyama2024VLFM}. 
After computing the access path, we navigate to the first frontier; as the environment updates during motion, we replan at each step and navigate to the new target via waypoint navigation.

\subsubsection{Waypoint Navigation Stage}
In the navigation stage, we perform A* path planning based on the target points and generate lookahead points at specific distances. We then adopt a point navigation scheme, utilizing the model trained with the Variable Experience Rollout (VER) reinforcement learning algorithm on the HM3D dataset \cite{2021HM3D} as a local planner, as described in \cite{yokoyama2024VLFM}. This planner uses depth and the relative position and orientation of the target points as inputs, producing the corresponding actions.

\section{EXPERIMENTAL RESULTS}
\begin{table*}[!t]  
    \centering
    \caption{Lifelong Multimodal Object Navigation Comparison.}
    \label{tab:result}
    \renewcommand{\arraystretch}{1.3} %
    \setlength{\tabcolsep}{8pt} %
    \begin{tabular}{ll c ccc ccc}
    \toprule
    & \multirow{2}{*}{\textbf{Method}}
    & \multirow{2}{*}{\textbf{Zero-shot}}
    & \multicolumn{3}{c}{\textbf{GOAT val\_seen}} 
    & \multicolumn{3}{c}{\textbf{GOAT val\_unseen}} \\
    \cmidrule(r){4-9}
    & 
    &
    & \makebox[0.9cm]{s-SR $(\uparrow)$} & \makebox[0.9cm]{e-SR$(\uparrow)$} & \makebox[0.9cm]{SPL$(\uparrow)$}
    & \makebox[0.9cm]{s-SR $(\uparrow)$} & \makebox[0.9cm]{e-SR$(\uparrow)$} & \makebox[0.9cm]{SPL$(\uparrow)$} \\
    \midrule
    & SenseAct-NN Skill Chain \cite{khanna2024goatbench} & $\times$  
        & 0.292 & -- & 0.128 & 0.295 & -- & 0.113 \\
    & SenseAct-NN Monolithic \cite{khanna2024goatbench} & $\times$ 
        & 0.168 & -- & 0.094 & 0.123 & -- & 0.068 \\
    & SenseAct-NN Monolithic\textsuperscript{†} & $\times$  
        & 0.159 & 0.003 & 0.104 & 0.126 & 0.008 & 0.088 \\
    \midrule
    & CoW \cite{gadre2023CoWs} & \checkmark & 0.148 & -- & 0.087 & 0.161 & -- & 0.104 \\ %
    & Modular GOAT \cite{khanna2024goatbench} & \checkmark & 0.263 & -- & \underline{0.175} & 0.249 & -- & \underline{0.172} \\ %
    & PIVOT \cite{nasiriany2024PIVOT} & \checkmark & -- & -- & -- & 0.102 & -- & 0.055 \\ %
    & VLMNav \cite{goetting2025EndtoEnd} & \checkmark & -- & -- & -- & 0.201 & -- & 0.096 \\ %
    & DyNaVLM \cite{ji2025DyNaVLM} & \checkmark & -- & -- & -- & 0.255 & -- & 0.102 \\ %
    & VLFM\textsuperscript{†} \cite{yokoyama2024VLFM} & \checkmark 
        & \underline{0.308} & \underline{0.025} & 0.129 & \underline{0.406} & \underline{0.028} & 0.159 \\
    & \textbf{SSMG-Nav(Ours)} & \checkmark & 
        \textbf{0.412} & \textbf{0.056} & \textbf{0.318} & \textbf{0.465} & \textbf{0.086} & \textbf{0.341} \\
    \bottomrule
    \end{tabular}
\end{table*}

\begin{table}[!t]  
    \centering
    \caption{Object Navigation Comparison.}
    \label{tab:result2}
    \renewcommand{\arraystretch}{1.2} %
    \setlength{\tabcolsep}{8pt} %
    \begin{tabular}{l c cc cc}
    \toprule
    \multirow{2}{*}{\textbf{Method}}
    & \multirow{2}{*}{\textbf{\makecell{Zero-\\shot}}}
    & \multicolumn{2}{c}{\textbf{HM3D}} 
    & \multicolumn{2}{c}{\textbf{MP3D}} \\
    \cmidrule{3-6}
    &
    & \makebox[0.6cm]{SR $(\uparrow)$} & \makebox[0.6cm]{SPL$(\uparrow)$}
    & \makebox[0.6cm]{SR $(\uparrow)$} & \makebox[0.6cm]{SPL$(\uparrow)$} \\
    \midrule
    PONI \cite{ramakrishnan2022PONI} & \texttimes %
        & -- & -- & 0.318 & 0.121 \\
    Habitat-Web \cite{ramrakhya2022HabitatWeb} & \texttimes %
        & 0.415 & 0.160 & 0.316 & 0.085 \\
    ZSON \cite{majumdar2022zson} & \texttimes %
        & 0.255 & 0.126 & 0.153 & 0.048 \\
    SGM \cite{zhang2024Imagine} & \texttimes %
        & \textbf{0.602} & 0.308 & 0.377 & 0.147 \\
    \midrule
    ESC \cite{zhou2023ESC} & \checkmark %
        & 0.392 & 0.223 & 0.287 & 0.142 \\
    L3MVN \cite{yu2023L3MVNa} & \checkmark %
        & 0.504 & 0.231 & 0.349 & 0.145 \\
    OpenFMNav \cite{kuang2024OpenFMNava} & \checkmark %
        & 0.549 & 0.244 & 0.372 & 0.157 \\
    SG-Nav \cite{yin2024SGNav} & \checkmark %
        & 0.540 & 0.249 & 0.402 & 0.160 \\
    VLFM \cite{yokoyama2024VLFM} & \checkmark 
        & 0.525 & 0.304 & 0.364 & \underline{0.175} \\
    VLFM\textsuperscript{†} & \checkmark 
        & 0.523 & 0.303 & 0.339 & 0.165 \\
    InstructNav \cite{long2024InstructNav} & \checkmark %
        & 0.510 & 0.187 & 0.420 & 0.161 \\
    UniGoal \cite{yin2025UniGoal} & \checkmark %
        & 0.545 & 0.251 & 0.410 & 0.164 \\
    WMNav \cite{nie2025WMNav} & \checkmark  %
        & 0.581 & \underline{0.312} & \textbf{0.454} & 0.172 \\
    \textbf{Ours} & \checkmark & 
        \underline{0.595} & \textbf{0.337} & \underline{0.443} & \textbf{0.191} \\
    \bottomrule
    \end{tabular}
\end{table}

\subsection{Experimental Setup}
\subsubsection{Benchmarks}
We evaluate on four high-resolution Habitat simulator-based benchmarks \cite{szot2021Habitat} spanning two tasks. First, we consider the challenging lifelong multimodal object navigation task introduced by GOAT-Bench \cite{khanna2024goatbench}. We use the \textit{GOAT val\_seen} and \textit{GOAT val\_unseen} splits; each split contains 36 indoor scenes, each scene includes 10 episodes, and each episode consists of 5--10 subtasks. Each subtask specifies a target object via one of three modalities: object class, a reference image, or a natural-language description.

To further assess our method's generality, we also evaluate on the classic object navigation task using two standard datasets, \textit{HM3D}\cite{2021HM3D} and \textit{MP3D}\cite{2017MP3D}. The first split comprises 2{,}000 episodes across 20 scenes with 6 goal categories, and the second split contains 2{,}195 episodes across 11 scenes with 21 goal categories. This task can be regarded as a special case of the lifelong multimodal setting---restricted to the single modality of object class---with exactly one subtask per episode.

\subsubsection{Metrics}
In object navigation tasks, the \textit{Success Rate (SR)} and \textit{Success Rate Weighted by Inverse Path Length (SPL)} are commonly used as evaluation metrics. 
Given the particularities of lifelong multimodal object navigation, we refine the overall SR into two granular metrics: \textit{subtask Success Rate (s-SR)} and \textit{episode Success Rate (e-SR)}. Concretely, s-SR is computed by first measuring the success of each subtask within an episode and then averaging these subtask-level success rates across all episodes. We deem an episode successful only if all of its subtasks are completed successfully; correspondingly, e-SR reports the proportion of episodes that satisfy this criterion, i.e., the fraction of fully completed episodes.

\subsubsection{Baselines}\label{subsubsec:baseline}
For lifelong multimodal object navigation, We compare SSMG-Nav against two RL methods, SenseAct-NN Skill Chain and Monolithic \cite{khanna2024goatbench}, and six zero-shot methods.
Among them, CoW \cite{gadre2023CoWs}, PIVOT \cite{nasiriany2024PIVOT}, and VLMNav \cite{goetting2025EndtoEnd} are memory-free (stateless) methods. Modular GOAT \cite{khanna2024goatbench} constructs an explicit semantic map, while DyNaVLM \cite{ji2025DyNaVLM} incorporates a graph-based memory architecture.
To establish a strong value-map baseline, we also adapt VLFM \cite{yokoyama2024VLFM} to the benchmark, carefully reconfiguring hyperparameters to ensure a fair comparison.

We evaluate four learning-based methods and eight zero-shot methods for the object-goal navigation task.
Among them, ESC \cite{zhou2023ESC}, L3MVN \cite{yu2023L3MVNa}, and OpenFMNav \cite{kuang2024OpenFMNava} are semantic map–based; VLFM \cite{yokoyama2024VLFM}, InstructNav \cite{long2024InstructNav}, and WMNav \cite{nie2025WMNav} are value map–based; and SG-Nav \cite{yin2024SGNav} and UniGoal \cite{yin2025UniGoal} are scene graph–based.

\subsubsection{Configuration Details}
The agent in Habitat is modeled as a cylinder with a radius of 0.17 m and a height of 1.44 m, following \cite{khanna2024goatbench}.
An RGB camera is mounted at a height of 1.31 m, with a resolution of 640 × 480 and a horizontal field of view  of 79\(^\circ\). 
A depth camera with the same intrinsics and extrinsics is also mounted, with a valid range of 0.5–5.0 m.
The action space is \{forward, turn left, turn right, stop\}; each forward action moves 0.25 m, and each rotation is 30\(^\circ\).
A task succeeds when the agent issues stop and is within 0.5 m of any candidate target viewpoint.
Each task is capped at 500 time steps.

\subsection{Benchmark Results}

\subsubsection{Performance on lifelong multimodal object navigation task}
The quantitative results are shown in \tref{tab:result}, where boldface and underlining indicate the best and second-best results, respectively.
The methods marked with \textsuperscript{†} (SenseAct-NN Monolithic and VLFM) represent our reimplementations, which include supplementary e-SR metrics to enhance analytical comprehensiveness.
Observing the results, VLFM achieves a strong SR, evidencing its exploration capability; however, its SPL remains low due to the lack of persistent memory.
In contrast, Modular GOAT benefits from its memory module and attains a substantially higher SPL than VLFM, albeit with a lower SR.
Our method attains the best performance across all metrics and clearly surpasses the baselines, with particularly large gains in SPL.
These results indicate that SSMG enables effective abstraction and retention of historical observations, allowing the agent to exploit past information during lifelong execution and leverage LLMs' general knowledge and reasoning for long-horizon decision-making, thereby achieving both high success and high efficiency.
On the other hand, the overall e-SR remains relatively low for all methods, underscoring that lifelong multimodal object navigation is still highly challenging and merits further investigation.

\begin{figure}[!t]
    \centering
    \subfloat{
        \includegraphics[width=0.8\linewidth]{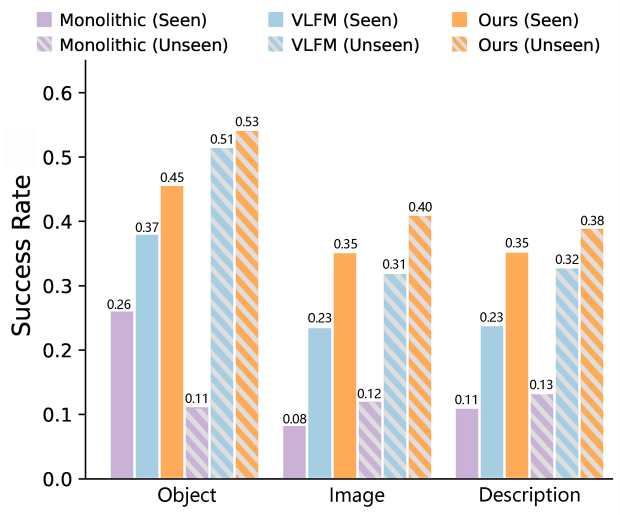}}
    \caption{Performance over different modalities.}
    \label{fig:modal}
\end{figure}

We further performed a statistical analysis of the performance of multimodal tasks, as shown in \fref{fig:modal}. Results demonstrate the superiority of our approach across all modalities. Notably, tasks utilizing the object modality generally underperform compared to those with image and description modalities, likely because instance-level object navigation requires a deeper understanding and analysis of target objects.  

\begin{figure}[!t]
    \centering
    \subfloat{
        \label{subfig:solver_mF1}
        \includegraphics[width=0.45\linewidth]{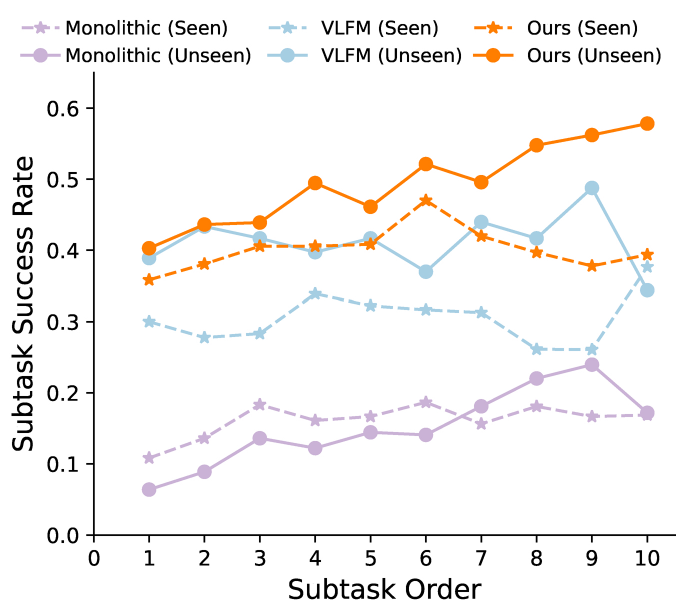}}
    \hfill
    \subfloat{
        \label{subfig:solver_AUC}
        \includegraphics[width=0.45\linewidth]{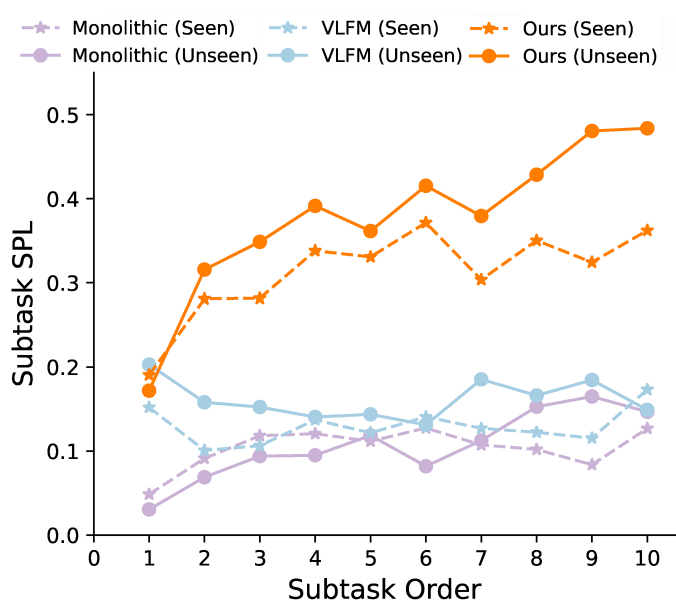}}
    
    \caption{Performance over time.}
    \label{fig:subtasks}
\end{figure}

To investigate the characteristics of different approaches on lifelong tasks, we compute performance statistics as a function of subtask index; the results are shown in ~\fref{fig:subtasks}. 
We first observe that our method consistently outperforms the baselines across almost all subtask orders, reinforcing the conclusions drawn from the previous figures and tables.
Moreover, both our method and the Monolithic baseline exhibit performance gains over subtasks (i.e., over time), relative to their initial states, with especially notable improvements in SPL on the \textit{GOAT val\_unseen} split for our method. 
As discussed in \sref{subsubsec:baseline}, the RL-based Monolithic approach implicitly encodes map information, whereas our approach explicitly abstracts historical observations into SSMG to assist decision making; thus, performance improvement over time is an expected and consistent finding. 
In contrast, VLFM outperforms Monolithic overall with general knowledge of VLMs, but it does not exhibit clear improvement over time, further underscoring the importance of memory in lifelong tasks, whether implemented explicitly or implicitly. 

\subsubsection{Performance on object navigation task}
The quantitative results in \tref{tab:result2} show that our method consistently outperforms in SPL while attaining comparable SR to strong baselines, unlike on GOAT-Bench.
We posit that the non-lifelong nature of this benchmark reduces the marginal benefit of persistent memory. We further attribute our SPL advantage primarily to the proposed long-horizon planner. Notably, the best-performing zero-shot method, WMNav, ingests a panoramic observation composed of six cameras at each step, whereas our approach relies solely on a single camera.

\subsection{Ablation Study}
\begin{table}
    \centering
    \caption{Ablation Study.}
    \label{tab:Ablation}
    \begin{tabular}{llccc}
        \toprule
        \multicolumn{2}{c}{\textbf{Method}} & \multirow{2}{*}{\textbf{s-SR}} & \multirow{2}{*}{\textbf{e-SR}} & \multirow{2}{*}{\textbf{SPL}}\\
        \cmidrule(r){1-2}
        Map & Planner & & & \\  %
        \midrule
        Semantic Map & Greedy & 0.389 & 0.056 & 0.278 \\
        Value Map & Greedy & 0.442 & \textbf{0.083} & 0.258 \\
        SSMG & Greedy & 0.453 & 0.056 & 0.328 \\
        SSMG & LHP w/o revisiting & 0.479 & \textbf{0.083} & 0.361 \\
        SSMG & LHP & \textbf{0.502} & \textbf{0.083} & \textbf{0.376} \\
        \bottomrule
    \end{tabular}
\end{table}

\begin{figure*}[!t]
    \centering
    \subfloat{
        \includegraphics[width=0.90\linewidth]{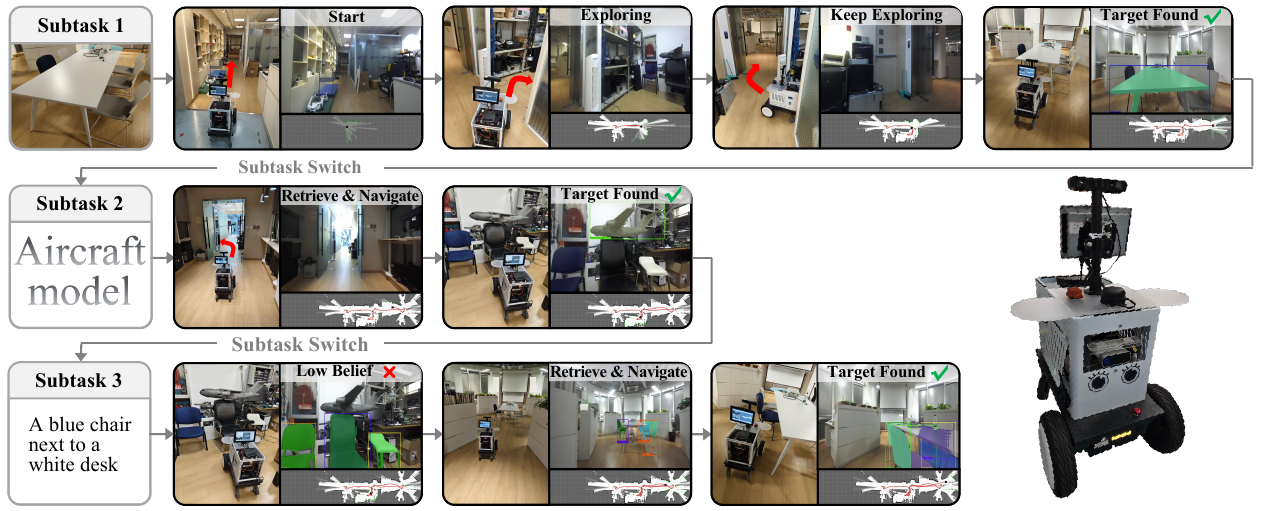}}
    \caption{Real-world demonstration of SSMG-Nav in an office environment with three subtasks.}
    \label{fig:real}
\end{figure*}

In the ablation study, we evaluate how map representations and planner strategies affect performance. 
Results on a downsampled subset of \textit{GOAT val\_unseen} are reported in \tref{tab:Ablation}.  
‘Semantic Map’ records only object semantics.
“Value Map” relies solely on the value map derived from the current target (i.e., no reusable long-term memory). 
‘Greedy’ selects the area with highest value or nearest distance.
Despite lower s-SR and e-SR than “Value Map,” the simplest “Semantic Map” method attains a higher SPL by leveraging memory, highlighting the importance of persistent memory for lifelong tasks.
For our method, combining ‘SSMG’ with ‘Greedy’ yields substantial gains over other map formats, demonstrating the advantage of SSMG as a memory representation.
Moreover, Performance improves with 'LHP w/o revisiting' and improves further with full 'LHP', indicating a progressive benefit and validating our module design.
The full version of our method achieves the best overall results.

\subsection{Real-world deployment}
Our method is deployed on a wheeled robotic platform. A ZED stereo camera provides RGB and depth data, and a 2D LiDAR supplies localization via Cartographer. We adopt \cite{zhang2020Falco} as the local planner to navigate to waypoints.
The experiment is conducted in an office environment with three subtasks: (i) an image-modality target specified as a “white desk,” (ii) an object-modality target specified as an aircraft model, and (iii) a text-modality target described as “a blue chair next to a white desk.” An overview of the process is shown in \fref{fig:real}.
This indicates that our method is able to perform lifelong multimodal object navigation on physical platforms.

\section{CONCLUSION}
In this paper, we introduced SSMG-Nav, a novel lifelong multimodal object navigation framework based on Semantic Skeleton Memory Graph (SSMG) that consolidates historical observations into a spatially grounded, reusable memory. By anchoring semantics to topological keypoints and clustering nearby entities into subgraphs, SSMG yields compact, decision-ready candidate destinations. Building on this representation, we synthesize multimodal prompts to elicit target beliefs from a VLM and plan with a long-horizon planner (LHP) that balances belief and traversability to minimize expected path length.
Experiments demonstrate that our method surpasses strong baselines on lifelong multimodal object navigation and standard object navigation, achieving substantial improvements in the challenging lifelong setting. We further confirm real-world feasibility.
However, several limitations remain.
First, the approach relies on large models, which introduces latency and constrains runtime speed in real-world deployments; as a result, the robot operates at relatively low velocities.
Second, the current method does not handle multi-floor scenarios. A promising direction is to extend SSMG with a voxel-based representation to build a hierarchical SSMG.

\bibliographystyle{IEEEtran}
\bibliography{reference}

@article{aghzal2025Survey,
  title={A survey on large language models for automated planning},
  author={Aghzal, Mohamed and Plaku, Erion and Stein, Gregory J and Yao, Ziyu},
  journal={arXiv preprint arXiv:2502.12435},
  year={2025}
}

@inproceedings{chang2023GOAT,
  title = {GOAT: GO to Any Thing},
  shorttitle = {GOAT},
  booktitle = {Robotics: Science and Systems},
  author = {Chang, Matthew and Gervet, Theophile and Khanna, Mukul and Yenamandra, Sriram and Shah, Dhruv and Min, So Yeon and Shah, Kavit and Paxton, Chris and Gupta, Saurabh and Batra, Dhruv and Mottaghi, Roozbeh and Malik, Jitendra and Chaplot, Devendra Singh},
  year = {2023},
  eprint = {2311.06430},
  archiveprefix = {arXiv},
  langid = {american}
}

@inproceedings{chen2024SpatialVLM,
  title = {SpatialVLM: Endowing Vision-Language Models with Spatial Reasoning Capabilities},
  shorttitle = {SpatialVLM},
  booktitle = {Proc. IEEE Conf. Comput. Vis. Pattern Recognit.},
  author = {Chen, Boyuan and Xu, Zhuo and Kirmani, Sean and Ichter, Brian and Sadigh, Dorsa and Guibas, Leonidas and Xia, Fei},
  year = {2024},
  month = jun,
  pages = {14455--14465},
  address = {Seattle, WA, USA},
  copyright = {https://doi.org/10.15223/policy-029},
  isbn = {979-8-3503-5300-6},
  langid = {american}
}

@article{croes1958Method,
  title={A method for solving traveling-salesman problems},
  author={Croes, Georges A},
  journal={Operations research},
  volume={6},
  number={6},
  pages={791--812},
  year={1958},
  publisher={INFORMS}
}

@inproceedings{gadre2023CoWs,
  title = {CoWs on Pasture: Baselines and Benchmarks for Language-Driven Zero-Shot Object Navigation},
  shorttitle = {CoWs on Pasture},
  booktitle = {Proc. IEEE Conf. Comput. Vis. Pattern Recognit.},
  author = {Gadre, Samir Yitzhak and Wortsman, Mitchell and Ilharco, Gabriel and Schmidt, Ludwig and Song, Shuran},
  year = {2023},
  month = jun,
  pages = {23171--23181},
  issn = {2575-7075},
  langid = {american}
}

@inproceedings{goetting2025EndtoEnd,
  title = {End-to-End Navigation with Vision-Language Models: Transforming Spatial Reasoning into Question-Answering},
  shorttitle = {End-to-End Navigation with Vision-Language Models},
  booktitle = {Proc. Adv. Neural Inform. Process. Syst.},
  author = {Goetting, Dylan and Singh, Himanshu Gaurav and Loquercio, Antonio},
  year = {2025},
  month = jul,
  pages = {22--35},
  issn = {2640-3498},
  langid = {english}
}

@inproceedings{gu2024ConceptGraphs,
  title = {ConceptGraphs: Open-Vocabulary 3D Scene Graphs for Perception and Planning},
  shorttitle = {ConceptGraphs},
  booktitle = {Proc. IEEE Int. Conf. Robot. Automat.},
  author = {Gu, Qiao and Kuwajerwala, Ali and Morin, Sacha and Jatavallabhula, Krishna Murthy and Sen, Bipasha and Agarwal, Aditya and Rivera, Corban and Paul, William and Ellis, Kirsty and Chellappa, Rama and Gan, Chuang and De Melo, Celso Miguel and Tenenbaum, Joshua B. and Torralba, Antonio and Shkurti, Florian and Paull, Liam},
  year = {2024},
  month = may,
  pages = {5021--5028},
  langid = {american}
}

@ARTICLE{ji2025DyNaVLM,
  title={DyNaVLM: Zero-Shot Vision-Language Navigation System with Dynamic Viewpoints and Self-Refining Graph Memory},
  author={Ji, Zihe and Lin, Huangxuan and Gao, Yue},
  journal={arXiv preprint arXiv:2506.15096},
  year={2025}
}

@inproceedings{khanna2024goatbench,
  title = {GOAT-bench: a benchmark for multi-modal lifelong navigation},
  booktitle = {Proc. IEEE Conf. Comput. Vis. Pattern Recognit.},
  author = {Khanna*, Mukul and Ramrakhya*, Ram and Chhablani, Gunjan and Yenamandra, Sriram and Gervet, Theophile and Chang, Matthew and Kira, Zsolt and Chaplot, Devendra Singh and Batra, Dhruv and Mottaghi, Roozbeh},
  year = {2024},
  langid = {american}
}

@inproceedings{kuang2024OpenFMNava,
  title = {OpenFMNav: Towards Open-Set Zero-Shot Object Navigation via Vision-Language Foundation Models},
  booktitle = {Findings Assoc. Comput. Linguist.: NAACL 2024},
  author = {Kuang, Yuxuan and Lin, Hai and Jiang, Meng},
  year = {2024},
  month = jun,
  pages = {338--351},
  address = {Mexico City, Mexico},
  langid = {american}
}

@inproceedings{li2023BLIP2,
  title={Blip-2: Bootstrapping language-image pre-training with frozen image encoders and large language models},
  author={Li, Junnan and Li, Dongxu and Savarese, Silvio and Hoi, Steven},
  booktitle={Proc. Int. Conf. Mach. Learn.},
  pages={19730--19742},
  year={2023},
}

@ARTICLE{long2024InstructNav,
       title={Instructnav: Zero-shot system for generic instruction navigation in unexplored environment},
  author={Long, Yuxing and Cai, Wenzhe and Wang, Hongcheng and Zhan, Guanqi and Dong, Hao},
  journal={arXiv preprint arXiv:2406.04882},
  year={2024}
}

@article{lu2025MultiView,
  title = {Multi-View Spatial Context and State Constraints for Object-Goal Navigation},
  author = {Lu, Chong and Liu, Meiqin and Luan, Zhirong and He, Yan and Chen, Badong},
  year = {2025},
  journal = {IEEE Robot. Automat. Lett.},
  pages = {1--8},
  issn = {2377-3766},
  langid = {american}
}

@inproceedings{majumdar2022zson,
  title={Zson: Zero-shot object-goal navigation using multimodal goal embeddings},
  author={Majumdar, Arjun and Aggarwal, Gunjan and Devnani, Bhavika and Hoffman, Judy and Batra, Dhruv},
  booktitle={Proc. Adv. Neural Inform. Process. Syst.},
  volume={35},
  pages={32340--32352},
  year={2022}
}

@inproceedings{nasiriany2024PIVOT,
author = {Nasiriany, Soroush and Xia, Fei and Yu, Wenhao and Xiao, Ted and Liang, Jacky and Dasgupta, Ishita and Xie, Annie and Driess, Danny and Wahid, Ayzaan and Xu, Zhuo and Vuong, Quan and Zhang, Tingnan and Lee, Tsang-Wei Edward and Lee, Kuang-Huei and Xu, Peng and Kirmani, Sean and Zhu, Yuke and Zeng, Andy and Hausman, Karol and Heess, Nicolas and Finn, Chelsea and Levine, Sergey and Ichter, Brian},
title = {PIVOT: iterative visual prompting elicits actionable knowledge for VLMs},
year = {2024},
booktitle = {Proc. Int. Conf. Mach. Learn.},
articleno = {1515},
numpages = {21},
location = {Vienna, Austria},
}

@inproceedings{nie2025WMNav,
  title={Wmnav: Integrating vision-language models into world models for object goal navigation},
  author={Nie, Dujun and Guo, Xianda and Duan, Yiqun and Zhang, Ruijun and Chen, Long},
  booktitle={Proc. IEEE Int. Conf. Intell. Robots Syst.},
  pages={2392--2399},
  year={2025},
}

@inproceedings{niuSkeletonBased,
  title={A skeleton-based topological planner for exploration in complex unknown environments},
  author={Niu, Haochen and Ji, Xingwu and Zhang, Lantao and Wen, Fei and Ying, Rendong and Liu, Peilin},
  booktitle={Proc. IEEE Int. Conf. Robot. Autom.},
  pages={11766--11772},
  year={2025},
}

@inproceedings{radford2021Learning,
  title={Learning transferable visual models from natural language supervision},
  author={Radford, Alec and Kim, Jong Wook and Hallacy, Chris and Ramesh, Aditya and Goh, Gabriel and Agarwal, Sandhini and Sastry, Girish and Askell, Amanda and Mishkin, Pamela and Clark, Jack and others},
  booktitle={Proc. Int. Conf. Mach. Learn.},
  pages={8748--8763},
  year={2021},
}

@inproceedings{rajvanshi2024SayNav,
  title = {SayNav: Grounding Large Language Models for Dynamic Planning to Navigation in New Environments},
  shorttitle = {SayNav},
  booktitle = {Proc. Int. Conf. Autom. Plan. Sched.},
  author = {Rajvanshi, Abhinav and Sikka, Karan and Lin, Xiao and Lee, Bhoram and Chiu, Han-pang and Velasquez, Alvaro},
  year = {2024},
  month = mar,
  langid = {american}
}

@inproceedings{ramakrishnan2022PONI,
  title = {PONI: Potential Functions for ObjectGoal Navigation with Interaction-free Learning},
  shorttitle = {PONI},
  booktitle = {Proc. IEEE Conf. Comput. Vis. Pattern Recognit.},
  author = {Ramakrishnan, Santhosh Kumar and Chaplot, Devendra Singh and {Al-Halah}, Ziad and Malik, Jitendra and Grauman, Kristen},
  year = {2022},
  month = jun,
  pages = {18868--18878},
  issn = {2575-7075},
  langid = {american}
}

@INPROCEEDINGS{ramrakhya2022HabitatWeb,
  author={Ramrakhya, Ram and Undersander, Eric and Batra, Dhruv and Das, Abhishek},
  booktitle={Proc. IEEE Conf. Comput. Vis. Pattern Recognit.}, 
  title={Habitat-Web: Learning Embodied Object-Search Strategies from Human Demonstrations at Scale}, 
  year={2022},
  volume={},
  number={},
  pages={5163-5173},
  doi={10.1109/CVPR52688.2022.00511}}

@inproceedings{szot2021Habitat,
  title={Habitat 2.0: Training home assistants to rearrange their habitat},
  author={Szot, Andrew and Clegg, Alexander and Undersander, Eric and Wijmans, Erik and Zhao, Yili and Turner, John and Maestre, Noah and Mukadam, Mustafa and Chaplot, Devendra Singh and Maksymets, Oleksandr and others},
  booktitle={Proc. Adv. Neural Inform. Process. Syst.},
  volume={34},
  pages={251--266},
  year={2021}
}

@article{tang2025OpenIN,
  title={Openin: Open-vocabulary instance-oriented navigation in dynamic domestic environments},
  author={Tang, Yujie and Wang, Meiling and Deng, Yinan and Zheng, Zibo and Deng, Jingchuan and Zuo, Sibo and Yue, Yufeng},
  journal={IEEE Robot. Automat. Lett.},
  volume={10},
  number={9},
  pages={9256--9263},
  year={2025},
  publisher={IEEE}
}

@inproceedings{wei2023ChainofThought,
  title={Chain-of-thought prompting elicits reasoning in large language models},
  author={Wei, Jason and Wang, Xuezhi and Schuurmans, Dale and Bosma, Maarten and Xia, Fei and Chi, Ed and Le, Quoc V and Zhou, Denny and others},
  booktitle={Proc. Adv. Neural Inform. Process. Syst.},
  volume={35},
  pages={24824--24837},
  year={2022}
}

@inproceedings{yin2024SGNav,
  title={Sg-nav: Online 3d scene graph prompting for llm-based zero-shot object navigation},
  author={Yin, Hang and Xu, Xiuwei and Wu, Zhenyu and Zhou, Jie and Lu, Jiwen},
  booktitle={Proc. Adv. Neural Inform. Process. Syst.},
  volume={37},
  pages={5285--5307},
  year={2024}
}

@inproceedings{yin2025UniGoal,
  title = {UniGoal: Towards Universal Zero-shot Goal-oriented Navigation},
  shorttitle = {UniGoal},
  booktitle = {Proc. IEEE Conf. Comput. Vis. Pattern Recognit.},
  author = {Yin, Hang and Xu, Xiuwei and Zhao, Lingqing and Wang, Ziwei and Zhou, Jie and Lu, Jiwen},
  year = {2025},
  eprint = {2503.10630},
  primaryclass = {cs},
  archiveprefix = {arXiv},
  langid = {american}
}

@inproceedings{yokoyama2024VLFM,
  title = {VLFM: Vision-Language Frontier Maps for Zero-Shot Semantic Navigation},
  shorttitle = {VLFM},
  booktitle = {Proc. IEEE Int. Conf. Robot. Automat.},
  author = {Yokoyama, Naoki and Ha, Sehoon and Batra, Dhruv and Wang, Jiuguang and Bucher, Bernadette},
  year = {2024},
  eprint = {2312.03275},
  archiveprefix = {arXiv},
  langid = {american}
}

@inproceedings{yu2023L3MVNa,
  title={L3mvn: Leveraging large language models for visual target navigation},
  author={Yu, Bangguo and Kasaei, Hamidreza and Cao, Ming},
  booktitle={Proc. IEEE Int. Conf. Intell. Robots Syst.},
  pages={3554--3560},
  year={2023},
}

@article{zhan2024MCGPT,
  title={Mc-gpt: Empowering vision-and-language navigation with memory map and reasoning chains},
  author={Zhan, Zhaohuan and Yu, Lisha and Yu, Sijie and Tan, Guang},
  journal={arXiv preprint arXiv:2405.10620},
  year={2024}
}

@article{zhang2020Falco,
  title = {Falco: Fast likelihood-based collision avoidance with extension to human-guided navigation},
  shorttitle = {Falco},
  author = {Zhang, Ji and Hu, Chen and Chadha, Rushat Gupta and Singh, Sanjiv},
  year = {2020},
  month = dec,
  journal = {Journal of Field Robotics},
  volume = {37},
  number = {8},
  pages = {1300--1313},
  issn = {1556-4959, 1556-4967},
  langid = {english}
}

@inproceedings{zhang2024FLTRNN,
  title={Fltrnn: Faithful long-horizon task planning for robotics with large language models},
  author={Zhang, Jiatao and Tang, Lanling and Song, Yufan and Meng, Qiwei and Qian, Haofu and Shao, Jun and Song, Wei and Zhu, Shiqiang and Gu, Jason},
  booktitle={Proc. IEEE Int. Conf. Robot. Autom.},
  pages={6680--6686},
  year={2024},
}

@inproceedings{zhang2024Imagine,
  title = {Imagine Before Go: Self-Supervised Generative Map for Object Goal Navigation},
  shorttitle = {Imagine Before Go},
  booktitle = {Proc. IEEE Conf. Comput. Vis. Pattern Recognit.},
  author = {Zhang, Sixian and Yu, Xinyao and Song, Xinhang and Wang, Xiaohan and Jiang, Shuqiang},
  year = {2024},
  month = jun,
  pages = {16414--16425},
  address = {Seattle, WA, USA},
  copyright = {https://doi.org/10.15223/policy-029},
  isbn = {979-8-3503-5300-6},
  langid = {english}
}

@inproceedings{zhao2024OVERNAV,
  title = {OVER-NAV: Elevating Iterative Vision-and-Language Navigation with Open-Vocabulary Detection and StructurEd Representation},
  shorttitle = {OVER-NAV},
  booktitle = {Proc. IEEE Conf. Comput. Vis. Pattern Recognit.},
  author = {Zhao, Ganlong and Li, Guanbin and Chen, Weikai and Yu, Yizhou},
  year = {2024},
  month = jun,
  pages = {16296--16306},
  address = {Seattle, WA, USA},
  copyright = {https://doi.org/10.15223/policy-029},
  isbn = {979-8-3503-5300-6},
  langid = {american}
}

@inproceedings{zhou2023ESC,
  title = {ESC: exploration with soft commonsense constraints for zero-shot object navigation},
  shorttitle = {ESC},
  booktitle = {Proc. Int. Conf. Mach. Learn.},
  author = {Zhou, Kaiwen and Zheng, Kaizhi and Pryor, Connor and Shen, Yilin and Jin, Hongxia and Getoor, Lise and Wang, Xin Eric},
  year = {2023},
  month = jul,
  volume = {202},
  pages = {42829--42842},
  address = {Honolulu, Hawaii, USA},
  langid = {american}
}

@article{Qwen-VL,
  title={Qwen-VL: A Versatile Vision-Language Model for Understanding, Localization, Text Reading, and Beyond},
  author={Bai, Jinze and Bai, Shuai and Yang, Shusheng and Wang, Shijie and Tan, Sinan and Wang, Peng and Lin, Junyang and Zhou, Chang and Zhou, Jingren},
  journal={arXiv preprint arXiv:2308.12966},
  year={2023}
}

@inproceedings{gdino,
title={Grounding dino: Marrying dino with grounded pre-training for open-set object detection},
  author={Liu, Shilong and Zeng, Zhaoyang and Ren, Tianhe and Li, Feng and Zhang, Hao and Yang, Jie and Jiang, Qing and Li, Chunyuan and Yang, Jianwei and Su, Hang and others},
  booktitle={Proc. Eur. Conf. Comput. Vis.},
  pages={38--55},
  year={2024},
  organization={Springer}
}

@article{mobile_sam,
  title={Faster segment anything: Towards lightweight sam for mobile applications},
  author={Zhang, Chaoning and Han, Dongshen and Qiao, Yu and Kim, Jung Uk and Bae, Sung-Ho and Lee, Seungkyu and Hong, Choong Seon},
  journal={arXiv preprint arXiv:2306.14289},
  year={2023}
}

@ARTICLE{2021HM3D,
  title={Habitat-matterport 3d dataset (hm3d): 1000 large-scale 3d environments for embodied ai},  author={Ramakrishnan, Santhosh K and Gokaslan, Aaron and Wijmans, Erik and Maksymets, Oleksandr and Clegg, Alex and Turner, John and Undersander, Eric and Galuba, Wojciech and Westbury, Andrew and Chang, Angel X and others},
  journal={arXiv preprint arXiv:2109.08238},
  year={2021}
}

@INPROCEEDINGS{2017MP3D,
  author={Chang, Angel and Dai, Angela and Funkhouser, Thomas and Halber, Maciej and Niebner, Matthias and Savva, Manolis and Song, Shuran and Zeng, Andy and Zhang, Yinda},
  booktitle={Proc. Int. Conf. 3D Vis.}, 
  title={Matterport3D: Learning from RGB-D Data in Indoor Environments}, 
  year={2017},
  volume={},
  number={},
  pages={667-676},
  doi={10.1109/3DV.2017.00081}}

\end{document}